\documentclass[10pt,twocolumn,letterpaper]{article}

\usepackage{iccv}
\usepackage{times}
\usepackage{epsfig}
\usepackage{graphicx}
\usepackage{amsmath}
\usepackage{amssymb}
\usepackage{graphicx}
\usepackage{amsmath}
\usepackage{amssymb}
\usepackage{booktabs}
\usepackage[usenames,dvipsnames]{xcolor}

\usepackage{floatrow}
\usepackage{comment}
\usepackage{amsthm}
\usepackage{color}
\usepackage{graphicx}
\usepackage{xspace}
\usepackage{multirow}
\usepackage{booktabs}
\usepackage{bbm}
\usepackage{array}
\usepackage{caption}
\usepackage{subcaption}
\usepackage{colortbl}
\usepackage{nicefrac}
\usepackage{url}
\usepackage{makecell}
\usepackage{hhline}
\usepackage{mathtools}

\usepackage[pagebackref=true,breaklinks=true,letterpaper=true,colorlinks,bookmarks=false]{hyperref}

\usepackage[capitalize]{cleveref}

\iccvfinalcopy

\ificcvfinal\fi

\newcommand{\figref}[1]{Fig.\xspace~\ref{#1}}
\newcommand{\tabref}[1]{Tab.\xspace~\ref{#1}}
\newcommand{\secref}[1]{Sec.\xspace~\ref{#1}}

\newcommand{\minisection}[1]{\noindent{\textbf{#1.}}}
\newcommand{\ApproachName}{TLC\xspace}
\newcommand{\ApproachNameLong}{\underline{T}oken-\underline{L}evel \underline{C}onfidence\xspace}

\newcommand{\OFALarge}{OFA$_{\text{Large}}$\xspace}
\newcommand{\OFABase}{OFA$_{\text{Base}}$\xspace}
\newcommand{\OFATiny}{OFA$_{\text{Tiny}}$\xspace}

\begin{document}

\title{Simple Token-Level Confidence Improves Caption Correctness}

 \author{Suzanne Petryk$^{1,2}$ \qquad
Spencer Whitehead$^2$ \qquad
Joseph E. Gonzalez$^1$ \qquad \\
Trevor Darrell$^1$ \qquad
Anna Rohrbach$^1$ \qquad
Marcus Rohrbach$^2$ \\ \\
\vspace{-.5em}
$^1$ UC Berkeley \qquad $^2$ Meta AI
\vspace{-.8em}
}

\maketitle
\ificcvfinal\thispagestyle{empty}\fi

\begin{abstract}

The ability to judge whether a caption correctly describes an image is a critical part of vision-language understanding. However, state-of-the-art models often misinterpret the correctness of fine-grained details, leading to errors in outputs such as hallucinating objects in generated captions or poor compositional reasoning. In this work, we explore \ApproachNameLong, or TLC, as a simple yet surprisingly effective method to assess caption correctness. Specifically, we fine-tune a vision-language model on image captioning, input an image and proposed caption to the model, and aggregate either algebraic or learned token confidences over words or sequences to estimate image-caption consistency. Compared to sequence-level scores from pretrained models, TLC with algebraic confidence measures achieves a relative improvement in accuracy by 10\% on verb understanding in SVO-Probes and outperforms prior state-of-the-art in image and group scores for compositional reasoning in Winoground by a relative 37\% and 9\%, respectively. When training data are available, a learned confidence estimator provides further improved performance, reducing object hallucination rates in MS~COCO Captions by a relative 30\% over the original model and setting a new state-of-the-art. 

\end{abstract}

\section{Introduction}
\label{sec:intro}

\begin{figure}[t]
  \centering
   \includegraphics[width=\linewidth]{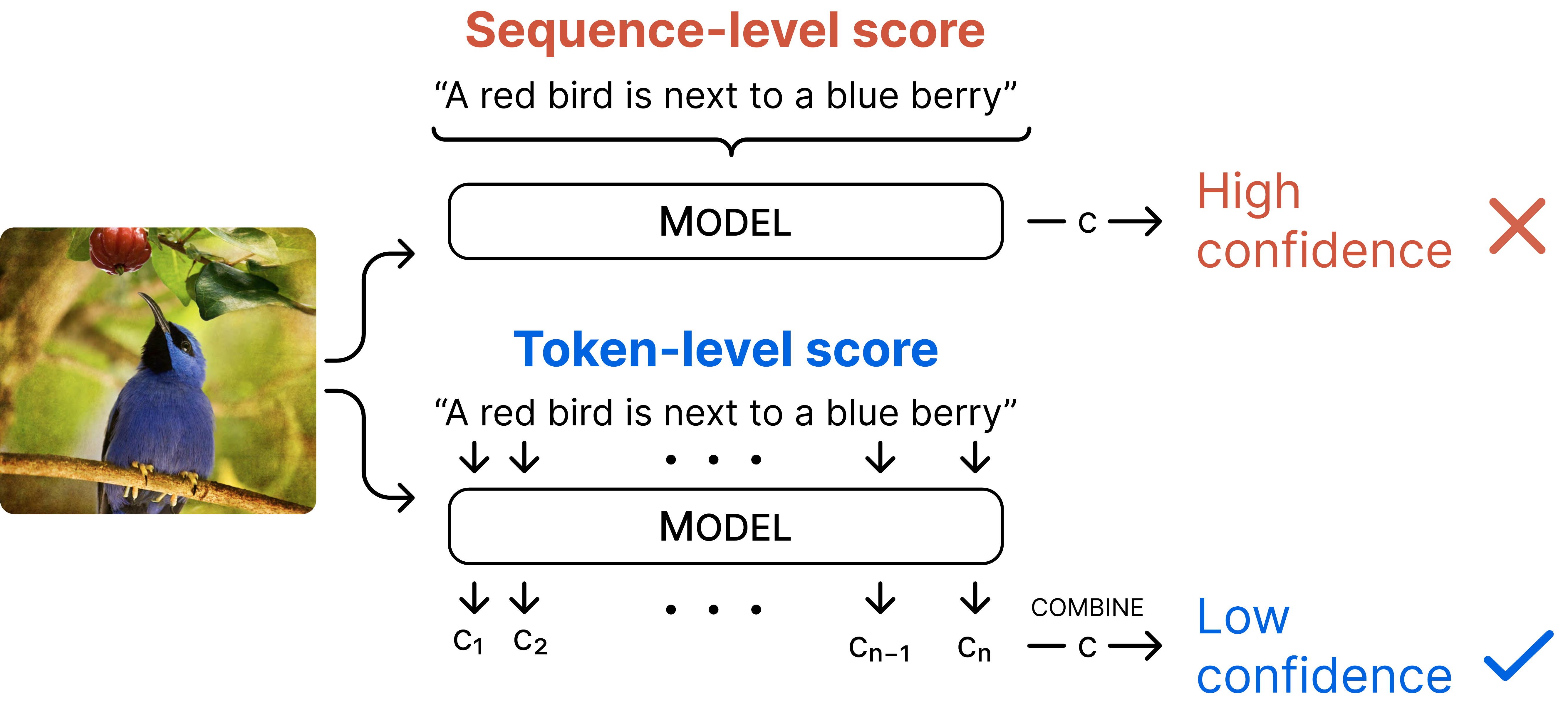}
   \caption{Judging caption correctness is still a challenge for large-scale models that operate at a sequence-level. We show that both algebraic and learned confidences at a token-level from a finetuned image captioning model improve fine-grained estimates of caption correctness.
   }
   \label{fig:teaser}
\end{figure}

For vision-and-language models, grounding and the ability to assess the correctness of a caption with respect to an image is critical for vision-language understanding.
When models have difficulties with these, the outputs can be error prone~\cite{rohrbach2018emnlp} or rely on biases~\cite{agrawal2018vqacp,hendricks2018women}.
State-of-the-art models, like CLIP~\cite{radford2021learning} or OFA~\cite{wang2022unifying}, demonstrate impressive capabilities in a variety of settings, in part, thanks to these properties.
While these models have had much success, recent efforts for probing state-of-the-art models have revealed some weaknesses in these areas.
For instance, the recent Winoground task~\cite{thrush_and_ross2022winoground} illustrates that these models, including large-scale pre-trained ones, can struggle to correctly associate image-caption pairs when the captions have differences in word order.
Similarly, SVO-Probes~\cite{hendricks2021probing} has shown that models can fail in situations that require understanding verbs compared to other parts of speech.
The observations from these probing tasks suggest that existing models have difficulties discerning fine-grained details that can appear in multimodal data.
This may hinder their accuracy and reliability when used in real settings, which presents significant issues in scenarios that require highly correct outputs, such as assisting people with visual impairments~\cite{gurari2018vizwiz,whitehead2022reliable}.

We conjecture that these weaknesses may be related to the granularity with which models perform image-text matching (ITM).
As shown in \cref{fig:teaser}, many existing models often operate at a sequence-level, pooling the representations of the image and caption to assess whether the text correctly describes the image.
This pretext task relies on sequence-level supervision and data with sufficient scale to learn finer-grained concepts, such as the difference between ``a cat jumping over a box'' and ``a box with a cat inside''.
Typical generative image captioning methods, on the other hand, generate words token-by-token and produce confidences for each one.
They are supervised at a token-level rather than sequence-level, which may emphasize the consistency of each token in a sequence more explicitly.

\begin{table}
  \centering
  \resizebox{1.\columnwidth}{!}{
  \begin{tabular}{l@{}ccc}
    \toprule
    \multirow{2}{*}{Dataset \& Task} & Winoground  & SVO-Probes & Hallucination in \\
                    & \cite{thrush_and_ross2022winoground} & \cite{hendricks2021probing} & Captioning \cite{rohrbach2018emnlp}\\[1ex]
     Metric  & Acc Image  $(\uparrow)$  &  Accuracy $(\uparrow)$ & \multicolumn{1}{c}{CHi $(\downarrow)$}\\
    \midrule
    Prior SoTA  & 19.75~\cite{ren2021learning} & -- & 3.2~\cite{li2022comprehending} \\
    Baseline (ours) & 10.25 & 81.23 &  2.0 \\
Ours & \textbf{27.00} & \textbf{89.47} &  \textbf{1.4} \\
\midrule
Rel. Improvement & 37\% & 10\% & 30\%\\
    \bottomrule
  \end{tabular}
  \caption{Summary of results. Despite its simplicity, the relative improvement over the next best approach highlights the significance of TLC for caption correctness.}
  \label{tab:resultsummary}
    \vspace{-2mm}
  }

\end{table}

Leveraging this observation, we explore \ApproachNameLong, or \ApproachName, for assessing image-caption correctness.
We input an image and proposed caption into a finetuned captioning model, which produces a distribution over the vocabulary at each time step. The base \ApproachName method, \ApproachName-A, uses algebraic confidence measures (\eg, softmax score) to compute confidence for a given token. To produce a single score for image-caption correctness, we either aggregate token confidences over the sequence (\eg, by taking the average value), or over particular words, such as verbs or objects. 
Next, we further investigate whether learned confidences can outperform algebraic ones.
We propose a \underline{L}earned confidence estimator, \ApproachName-L, for use in the caption generation setting where training data is available. We use existing annotations to model the likelihood that a predicted token matches reference tokens, and an additional validation set to calibrate our estimated confidence to actual correct and incorrect concepts. Using \ApproachName-L to re-rank candidate captions, we reduce hallucination rates in the final output captions.

Both \ApproachName-A and \ApproachName-L are simple to implement and can be applied on top of any autoregressive image captioning model with an encoder and decoder, an architecture found to scale well with data and multimodal tasks~\cite{chen2022pali,ofa,wang2021simvlm,wang2022git}.
In this work, we demonstrate the effectiveness of token-level confidence across multiple model sizes of OFA~\cite{ofa}, a recent Transformer-based model~\cite{vaswani2017attention} with strong performance on many vision-language tasks. As summarized in \tabref{tab:resultsummary}, on the challenging Winoground~\cite{thrush_and_ross2022winoground} benchmark evaluating compositional reasoning, we show that \ApproachName-A more than doubles accuracy over pretrained ITM scores, \eg, from 10.25\% to 27\% on image score (\secref{sec:experiment-winoground}). \ApproachName-A additionally shows a relative improvement of image and group scores of 37\% and 9\%, respectively, over the prior state-of-the-art on Winoground~\cite{ren2021learning}, which used a regularization tailored for multimodal alignment. \ApproachName-A also outperforms ITM on a fine-grained verb understanding task~\cite{hendricks2021probing} by a relative 10\% (\secref{sec:experiment-svo}). When using \ApproachName-L to re-rank candidate captions on MS~COCO~\cite{chen2015microsoft}, we achieve a 30\% relative reduction in object hallucination rate over the original captions and set a new state-of-the-art on a hallucination benchmark~\cite{rohrbach2018emnlp} (\secref{sec:experiment-coco}).
These results demonstrate that token-level confidence, whether algebraic (\ApproachName-A) or learned (\ApproachName-L) are a powerful yet simple resource for improving multimodal reliability.

\section{Related Work}
\label{sec:relatedwork}

\minisection{Caption correctness}
One of the desired properties of a good caption is correctness, \textit{i.e.}, being faithful to an image. \cite{hendricks2021probing,park2022exposing,thrush_and_ross2022winoground} propose benchmarks to probe for sensitivity to hard negatives of different types, such as compositional reasoning or action understanding. We use probing benchmarks in our work to demonstrate the effectiveness of \ApproachName-A.
Within caption generation, \cite{rohrbach2018emnlp} notes that in practice, image captioning models suffer from object hallucination~\cite{rohrbach2018emnlp}, driven by visual misclassification and over-reliance on language priors. Several recent works addressed the issue of object hallucination~\cite{biten2022let,li2022comprehending}, in some cases relying on causal inference-based approaches~\cite{liu2022show,yang2021causal,yang2021deconfounded}. Other recent works pose a slightly distinct problem of correcting errors in a caption provided for a given image (i.e., not as part of the caption generation process)~\cite{Sammani2019ModificationNet,sammani2020show,wang2022explicit}. 
Some works propose caption decoding methods such as constrained beam search~\cite{anderson2016guided}, an uncertainty-aware beam search using prediction entropy~\cite{xiao2021hallucination}, or a non-autoregressive caption decoding method~\cite{fei2022efficient} to target criteria such as correctness. However, the original formulation of beam search remains the dominant decoding method used in modern multimodal architectures~\cite{chen2022pali,li2023blip,ofa,wang2022image}. We apply our approach on top of captions generated with beam search and demonstrate that simply re-ranking beams based on token confidences can reduce hallucinations.

\minisection{Correctness estimation in language models}
Similar issues around correctness and hallucination are also relevant for many language-only tasks that require autoregressive prediction. Hallucination in particular has been studied for tasks like abstractive summarization~\cite{maynez-etal-2020-faithfulness}, e.g., one work performs token-level hallucination detection~\cite{zhou2020detecting}. A number of works study model uncertainty and aim to improve model calibration for machine translation~\cite{glushkova2021uncertainty,guerreiro2022looking,wang-etal-2020-inference}, dialog~\cite{mielke2022reducing}, question answering~\cite{zhang2021knowing} and spoken language understanding~\cite{shen2020modeling}, to name a few tasks.
While our focus on image captioning is similarly a conditional generation task, estimating confidence in the multimodal setting can be challenging as errors are driven by factors from both modalities~\cite{whitehead2022reliable}.

\minisection{Image captioning}
Image captioning has seen significant progress since the arrival of deep learning as a dominant methodology~\cite{anderson2018bottom,donahue2015long,huang2019attention,karpathy2015deep,rennie2017self,vinyals2015show}. In recent years Transformer-based architectures have gained particular prominence~\cite{li2020oscar,shen2021much,zhang2021vinvl}. Many papers take the approach of pretraining large vision-and-language models and then adapting them to downstream tasks, including captioning~\cite{li2020unimo,wang2021simvlm}. Recent efforts focus on further scaling these pretraining-based methods~\cite{alayrac2022flamingo,hu2022scaling,wang2022git,yu2022coca}, while many also aim to unify multiple vision-and-language tasks during pretraining~\cite{chen2022pali,cho2021unifying,ofa,wang2022image}. Despite steady improvements in image caption quality over the past years, even the best models still make mistakes. Here, we study the reliability of vision-language models, with the goal of assessing caption correctness.

\minisection{Reliability in multimodal models}
With the adoption of Large Language/Vision/Vision-and-Language Models (LLMs, LVMs, LVLMs), it is increasingly important to study their limitations and outline expectations regarding their \emph{reliability}. One of the first efforts in doing that for LLMs and LVMs (unimodally) is \cite{tran2022plex}, whose broad definition of reliability includes aspects from modeling uncertainty to robust generalization and adaptation. 
A recent work in multimodal learning outlines reliability of visual question answering~\cite{whitehead2022reliable}, defining it as a model's ability to ensure a low risk of error by means of abstaining from answering. In our work, we approach reliability by improving assessments of caption correctness, and incorporating these estimates to reduce rates of error in generated captions.

\section{\ApproachName: Token-Level Confidence for Caption Correctness}
\label{sec:method}

\minisection{Overview} Given an image and a caption, \ApproachName produces a confidence score for each token and aggregates these scores to produce an estimate of caption correctness, \textit{i.e.}, semantic consistency with the image. First, we describe two forms of confidences: algebraic (\ApproachName-A, \secref{sec:method-algebraic}) and learned (\ApproachName-L, \secref{sec:method-learning}). Next, in \secref{sec:method-using-confidences}, we describe how to combine token confidences to measure caption correctness and use token confidences to re-rank captions during generation. In our experiments, we will then verify \ApproachName-A primarily on out-of-domain probing benchmarks (Sections \ref{sec:experiment-winoground} and \ref{sec:experiment-svo}). We then evaluate \ApproachName-L in a setting where in-domain training data is available (\secref{sec:experiment-coco}).

\minisection{Preliminaries} Let $f_{\mathit{pre}}$ be a vision-language model pretrained on a large multimodal dataset, and $f_{\mathit{cap}}$ be a model initialized with $f_{\mathit{pre}}$ and subsequently finetuned for autoregressive image captioning.
Given an image $x$, a caption consists of a sequence of $n$ tokens $t_{1:n}$ describing the image. At each decoding time step $k \in \{1...n\}$, $f_{\mathit{cap}}$ produces a distribution of token likelihoods $\vec{z}_k \in \mathbb{R}^{|V|}$ for a vocabulary $V$, conditioned on previous outputs $\vec{z}_{1:k-1}$. Autoregressive captioning models are typically trained with a token-level cross-entropy loss on $\vec{z}_k$, often followed by self-critical sequence training~\cite{rennie2017self}. Decoding methods such as sampling or beam search can then be used to select tokens at inference time, typically aiming to maximize the image-conditional sequence likelihood.

\subsection{\ApproachName-A: Algebraic Confidences}
\label{sec:method-algebraic}

A simple method for measuring token-level confidence is to use an algebraic function of the distribution $\vec{z}_k$ directly, such as taking the logit or softmax value at the selected token index. We refer to token confidences derived from algebraic functions of $\vec{z}_k$ as \ApproachName-\underline{A}lgebraic, or \ApproachName-A. Prior works find simple measures such as softmax to be unreliable in both vision and vision-language ``one-of-K'' classification tasks~\cite{guo2017calibration,whitehead2022reliable}. In contrast, we find that softmax scores from autoregressively-generated tokens perform surprisingly well, even on data that is out-of-distribution from the image captioning training set used by $f_{\mathit{cap}}$. This is aligned with findings in the language-only setting~\cite{desai2020calibration,stengel2022calibrated,varshney2022investigating}, suggesting that token-level language modeling may be key for reliable confidence measures.

\subsection{\ApproachName-L: Learned Domain-Specific Confidences}
\label{sec:method-learning}

\begin{figure*}
  \centering
     \includegraphics[width=\linewidth]{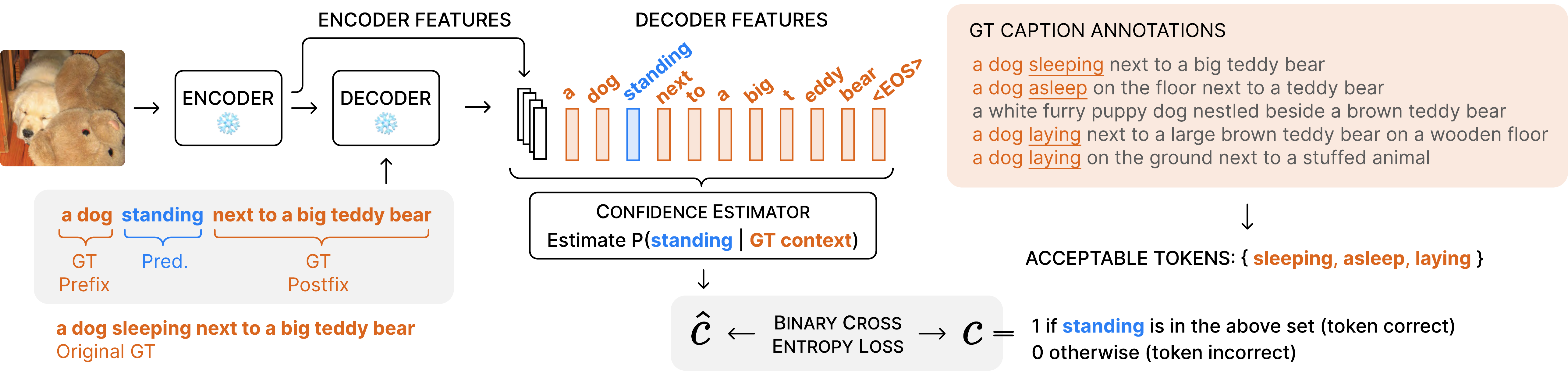}
  \caption{\ApproachName-L: A framework to learn token-level confidence for a pretrained autoregressive encoder-decoder captioning model. We first use the captioning model to predict the next token (\eg, ``standing'') after a partial reference caption (\eg, ``a dog''), shown in the bottom left. We input this sequence along with the image and the rest of the reference caption to the model, and obtain corresponding encoder and decoder features. These features become the inputs to our confidence estimator, a Transformer encoder. For supervising correctness, we create a binary classification task to learn whether or not the model's predicted token matched any reference token at the same time step with the same prefix.
  }
  \label{fig:method}
\end{figure*}

Although we observe that \ApproachName-A performs well on evaluation benchmarks out-of-distribution from the image captioning training data (Sections \ref{sec:experiment-winoground} and \ref{sec:experiment-svo}), we would like to see whether \textit{learning} a confidence estimator on in-distribution training data could improve estimates of correctness, similar to~\cite{whitehead2022reliable}.
However, we do not have direct supervision to measure the correctness of a specific token in an arbitrary predicted caption with an image, aside from human evaluation. Instead, we leverage existing reference captions to learn a binary classification task, measuring whether a predicted token matches one or more reference tokens at the same time step. \figref{fig:method} presents an overview of this method, which we refer to as \ApproachName-\underline{L}earned, or \ApproachName-L.

\minisection{Forming the training set} We begin with a trained and frozen $f_{\mathit{cap}}$ and use a heldout dataset $\mathcal{X}$ for training a confidence estimator $g$. Compared to the training set for $f_{\mathit{cap}}$, $\mathcal{X}$ provides a better estimate of the captioning model's performance on test data. In this work, we simply use the $f_{\mathit{cap}}$ validation set. For each image in $\mathcal{X}$, paired with one or more references, we select one of the reference captions $t_{1:n}$ and time step $k$ within the caption. We first input the \textit{prefix}, or $t_{1:k-1}$, into the $f_{\mathit{cap}}$ decoder to predict the next token, $\hat t_k$. We assign a binary label $c$ to $\hat t_k$ -- it is \textbf{correct} $(c=1)$ if it matches the reference token $t_k$ or any token at $k$ from other reference captions with the same prefix. Otherwise, $\hat t_k$ is labeled as \textbf{incorrect} $(c=0)$. For example, in~\figref{fig:method}, the original reference token $t_k$ is ``sleeping'', yet ``asleep'' and ``laying'' are also considered correct, given that they share the same prefix ``a dog''. The predicted token $\hat t_k$ ``standing'' is therefore labeled as incorrect. This provides proxy for true consistency with the image, which may be noisy; for example, ``resting'' would be considered incorrect in~\figref{fig:method}. Nevertheless, these labels enable \ApproachName-L to learn effective in-domain confidences (\secref{sec:experiment-coco}). At each epoch, we re-sample a reference caption and a time step $k$ for each image in order to leverage all available ground-truth tokens.

\minisection{Training a confidence estimator} The output of $g$ is a scalar $\hat c$, trained with binary cross-entropy loss with $c$ as supervision. As input, $g$ receives image features from the model, such as those output by an encoder. It also receives token-level features from the decoder (\eg, just before decoder features are projected into the vocabulary space). We find that including the reference \textit{postfix}, or $t_{k+1:n}$, in addition to the prefix $t_{1:k}$ and predicted token $\hat t_k$ improves the confidence estimation. We pass the encoder features and position-encoded decoder sequence into a Transformer encoder~\cite{vaswani2017attention}, and pass the output embedding of token $\hat t_k$ into a small feed-forward network to produce $\hat c$. We provide details on our specific choice of architecture in~\secref{sec:experiment-setup}. At inference time, we run our confidence estimator once per time step within a predicted caption $\hat t_{1:n}$.

\minisection{A bidirectional confidence} Although we supervise confidence for a single token $\hat t_k$ at a time, the full caption context is given as input. Due to self-attention in the Transformer encoder within $g$, the final prediction $\hat c$ represents a bidirectional confidence estimate, in contrast to the original autoregressive token predictions. This enables a useful combination: generating tokens autoregressively scales well with data and model size~\cite{ofa,chen2022pali}, whereas estimating token confidence bidirectionally uses future context to inform correctness.

\subsection{From Confidence to Caption Correctness}
\label{sec:method-using-confidences}

\subsubsection{Combining Confidences}
\label{sec:method-combining}

In practice, we would like to measure correctness over an entire caption or particular span, such as a word or phrase. To obtain such a score from token-level confidences, we can simply aggregate the confidences over a specific span of tokens $t_{i:j}$ or the full sequence $t_{1:n}$ by taking, \eg, the minimum or average confidence value. We exclude the end-of-sentence (EOS) token, as its confidence is often poorly calibrated relative to previous tokens~\cite{kumar2019calibration}. In our experiments, we compare correctness between image-caption pairs by aggregating over the full sequence (\secref{sec:experiment-winoground}) or specific words (Sections \ref{sec:experiment-svo} and \ref{sec:experiment-coco}).

\subsubsection{Confidence During Caption Generation}
\label{sec:method-reducing-hallucinations}

We can use token-level confidences to not only estimate correctness between an image and an \textit{existing} caption but also between a \textit{proposed} caption candidate during generation. By re-ranking candidates relative to estimated correctness, we can reduce errors in the final selected captions.

When generating a caption, it is common to first predict a set of $B$ candidate captions using an autoregressive decoding method such as beam search. Initially, the beams are ranked according to their cumulative token log likelihoods from the captioning model:

\begin{equation}
\mathbb{P}(t_{1:n}) = \sum_{k=1}^{n} \text{log}\, p(t_k \mid t_{1:k-1}, x)
    \label{eq:beam-ranks}
\end{equation}

However, token likelihood can fail to rank captions that are fully correct above those that contain an error. For example, a fluent and detailed sentence with a single-word hallucination may rank above a simpler, yet correct, caption. This is observed in~\cite{rohrbach2018emnlp}, where captions with higher CIDEr~\cite{vedantam2015cider} could also have higher hallucination rates. It is also similar to prior work in machine translation~\cite{guerreiro2022looking}, which noted that errors can be ``bad luck'' from generation rather than inherent model failure.

To alleviate this, we first define a set of words or concepts $\mathcal{S}$ that we estimate correctness for. For example, in our experiments, we consider only the tokens that correspond to MS~COCO~\cite{chen2015microsoft} object categories, as we have annotations for their correctness during validation and evaluation. Beginning from the highest-likelihood beam, we estimate confidence $\hat c$ for each set of words in $\mathcal{S}$ that appear in the beam (\eg, each MS~COCO object that is mentioned). If any $\hat c$ are less than a threshold $\gamma$, we reject the beam, and continue to the next one until we reach a beam where all relevant tokens are predicted to be correct ($\hat c \geq \gamma$), or where there are no tokens from $\mathcal{S}$. If none of the beams satisfy these criteria, we output the original (highest-likelihood) caption. In that setting, we could alternatively choose to abstain from providing a caption in order to avoid misleading a user, similar to~\cite{whitehead2022reliable}. However, we instead choose the original caption in our experiments to simplify the comparison between methods.

We choose the threshold $\gamma$ on a validation set to control the rate of false positives. This is captured by the precision: \textit{``out of all samples predicted as correct, what fraction are} actually \textit{correct?''} We define a target precision $\alpha$, such as 99\%, and select $\gamma$ such that the binary decisions $\hat c \geq \gamma$ maximize the recall of correct samples in $\mathcal{S}$ on the validation set.

\section{Experiments}
\label{sec:experiments}

After discussing the experimental setup (\secref{sec:experiment-setup}), we demonstrate the effectiveness of \ApproachName-A for identifying correct image-caption pairs that test understanding of compositionality (\secref{sec:experiment-winoground}) and verbs (\secref{sec:experiment-svo}). We then evaluate both \ApproachName-A and \ApproachName-L on reducing object hallucinations in generated captions (\secref{sec:experiment-coco}).

\subsection{Experimental Setup}
\label{sec:experiment-setup}

As a captioning model, we choose to experiment with OFA~\cite{ofa}, a recent open-source sequence-to-sequence multimodal transformer that achieves state-of-the-art captioning performance. OFA has a simple encoder-decoder architecture designed to unify multimodal tasks conditioned on an image and specific input instruction (e.g., ``What does the image describe?'' prompts the model to output a sequence of tokens for captioning). We use the official implementation and checkpoints ($f_{\mathit{pre}}$) for \OFALarge, \OFABase, and \OFATiny, pretrained on a dataset with 20M publicly available image-text pairs. As image-text matching was included as a task in OFA pretraining, we use \textbf{ITM} in our results to denote the image-text matching score from $f_{\mathit{pre}}$. For $f_{\mathit{cap}}$, we finetune each scale of OFA model on MS~COCO Captions~\cite{chen2015microsoft}, which has about 80k training images. We split the validation set of 40k images into three parts for training, validation, and testing of $g$, following~\cite{whitehead2022reliable}. Additional dataset details are in Appendix \ref{sec:supp-dataset}.

For \ApproachName-A, we use the softmax score at the selected token index. We experiment with several other choices of algebraic function and report results in Appendix \ref{sec:supp-precision-recall}. For \ApproachName-L, as input to the learned confidence estimator $g$, we use multimodal image and instruction features output from the OFA encoder, as well as token embeddings from the decoder just before they are projected onto the logit space by a linear layer. $g$ itself is a 4-layer Transformer encoder~\cite{vaswani2017attention}, followed by a 2-layer MLP. We add a learned positional encoding to the token features, and train $g$ for 200 epochs on 8 V100 GPUs. Additional details are in Appendix~\ref{sec:supp-model}.

\subsection{Correctness Around Compositional Reasoning}
\label{sec:experiment-winoground}

\begin{table}
  \centering
  \resizebox{1.\columnwidth}{!}{
  \begin{tabular}{@{}llrrrr}
    \toprule
    Model & Conf. & Text & Image & Group \\
    \midrule
    MTurk Human~\cite{thrush_and_ross2022winoground} & -  & 89.50 & 88.50 & 85.50 \\
    Random Chance~\cite{thrush_and_ross2022winoground} & - & 25.00 & 25.00 & 16.67 \\
    \midrule
    UNITER$_{\text{Large}}$~\cite{thrush_and_ross2022winoground} & ITM & 38.00 & 14.00 & 10.50 \\
    VinVL~\cite{thrush_and_ross2022winoground} & ITM & 37.75 & 17.75 & 14.50 \\
    CACR$_{\text{Base}}$~\cite{pandey2022cross} & CACR & 39.25 & 17.75 & 14.25 \\
    IAIS$_{\text{Large}}$~\cite{pandey2022cross} & IAIS & \textbf{*}42.50 & 19.75 & 16.00 \\
    \midrule
   \multirow{3}{*}{\OFALarge} & ITM & \textbf{30.75} & 10.25 & 7.25 \\
      &  \ApproachName-A & 29.25 & \textbf{*27.00} & \textbf{*17.50} \\
      &  $(\Delta)$ & \textcolor{Red}{($-$1.5)}  & \textcolor{ForestGreen}{($+$16.75)} &  \textcolor{ForestGreen}{($+$10.25)} \\
      \midrule
    \multirow{3}{*}{\OFABase} & ITM & \textbf{26.75} & 10.75 & 6.50 \\
     & \ApproachName-A &  24.50 & \textbf{23.50} & \textbf{13.75}\\
     &  $(\Delta)$  & \textcolor{Red}{($-$2.25)}  & \textcolor{ForestGreen}{($+$12.75)} &  \textcolor{ForestGreen}{($+$7.25)} \\
    \midrule
    \multirow{3}{*}{\OFATiny}  &  ITM & \textbf{22.75} & 7.75 & 4.50 \\
     & \ApproachName-A & 16.50 & \textbf{15.75} & \textbf{6.75} \\
        &  $(\Delta)$  & \textcolor{Red}{($-$6.25)}  & \textcolor{ForestGreen}{($+$8.00)} &  \textcolor{ForestGreen}{($+$2.25)} \\
    \bottomrule
  \end{tabular}
  \caption{Accuracy on text, image, and group score for the Winoground evaluation dataset~\cite{thrush_and_ross2022winoground}.
  Citations indicate where scores are reported, and \textbf{*} indicates state-of-the-art.
  }
  \label{tab:winoground}}
\end{table}

First, we assess the ability of \ApproachName-A to select corresponding image-caption pairs. We use Winoground~\cite{thrush_and_ross2022winoground}, a dataset curated to test the compositionality of vision-language models. Each of the 400 examples contains two image-caption pairs $(I_0, C_0) $ and $(I_1, C_1)$. Captions $C_0$ and $C_1$ contain the same words and/or morphemes, yet differ in order; for example, ``there is a mug in some grass'' and ``there is some grass in a mug''. There are three evaluations per example: text score (given an image, select the correct caption), image score (given a caption, select the correct image), and group score (all text and image scores for an example must be correct). A pairing is considered correct if the image-caption matching score for the correct pair is greater than that of the incorrect pair (\textit{i.e.}, $c_{\mathit{POS}} > c_{\mathit{NEG}}$). \cite{thrush_and_ross2022winoground} find that the task is surprisingly difficult, with all models they test performing below random chance for image and group score.

As correctness estimates, \cite{thrush_and_ross2022winoground} use image-text matching scores (ITM) from a range of pretrained vision-language models. Other works~\cite{pandey2022cross,ren2021learning} design training losses specifically targeting relation alignment. Using \ApproachName-A, we produce a correctness estimate $c$ by simply averaging token-level softmax scores for each proposed image-caption pair. We present results in~\tabref{tab:winoground}.

\minisection{\ApproachName-A outperforms prior SOTA image and group performance} \ApproachName-A with \OFALarge reaches above random chance for both image and group score, improving over prior state-of-the-art. Despite its simplicity, with no additional training beyond standard image captioning, \ApproachName-A outperforms IAIS (proposed in \cite{ren2021learning}), a training method optimized for multimodal attention alignment. Compared to ITM across OFA model sizes, \ApproachName-A more than doubles the image and group scores in all but one case (\OFATiny group).

\subsection{Correctness Around Verb Understanding}
\label{sec:experiment-svo}

\begin{table}
  \centering
    \resizebox{0.8\columnwidth}{!}{
  \begin{tabular}{@{}lrrr}
    \toprule
    \multirow{2}{*}{Confidence} & \multicolumn{3}{c}{Model} \\ \cline{2-4}
      \rule{0pt}{2.5ex}  &  \OFALarge & \OFABase & \OFATiny \\
     \midrule
     ITM & 81.23 & 78.44 & 65.25 \\
     \ApproachName-A & \textbf{89.47} & \textbf{89.64} & \textbf{81.34}  \\
    $(\Delta)$  & \textcolor{ForestGreen}{($+$8.24)}  & \textcolor{ForestGreen}{($+$11.20)} &  \textcolor{ForestGreen}{($+$16.09)} \\
    \bottomrule
  \end{tabular}
  \caption{Image-caption matching accuracy for verb understanding with a subset of SVO-Probes~\cite{hendricks2021probing}. \ApproachName-A uses token-level softmax scores aggregated over the verb in each example.}
  \label{tab:svo}}
\end{table}

Next, we consider caption correctness when aggregating token confidences over a single word, rather than over a full sequence as in~\secref{sec:experiment-winoground}. To evaluate this, we use SVO-Probes, a dataset designed by Hendricks and Nematzadeh~\cite{hendricks2021probing} to test the verb understanding of vision-language transformers. Each example contains an image and a caption describing a $\langle$subject, verb, object$\rangle$ relation in the scene. It also contains a negative image, where only one part of the relation is different, such as $\langle$person, swim, water$\rangle$ and  $\langle$person, walk, water$\rangle$.
We use a publicly available subset of about 6,500 examples for verb understanding, and use a parser~\cite{spacy} to annotate the location of the verb in each caption. We aggregate token confidences over the verb tokens for \ApproachName-A. \tabref{tab:svo} presents image-caption accuracy, where a score is 1 if the confidence is greater for the correct image (again, if $c_{\mathit{POS}} > c_{\mathit{NEG}}$).

\minisection{\ApproachName-A outperforms image-text matching scores} From~\tabref{tab:svo}, we see that \ApproachName-A reaches higher image-caption matching accuracy compared to the ITM scores from pretrained models, across a range of model sizes (\eg, 8.24\% and 11.20\% improvement for \OFALarge and \OFABase respectively). Therefore, when localized word or token positions are available, they can be leveraged for a finer-grained matching score than ITM operating on the full sequence.

\subsection{Reducing Object Hallucinations}
\label{sec:experiment-coco}

\begin{figure*}
  \centering
   \includegraphics[width=0.95\linewidth]{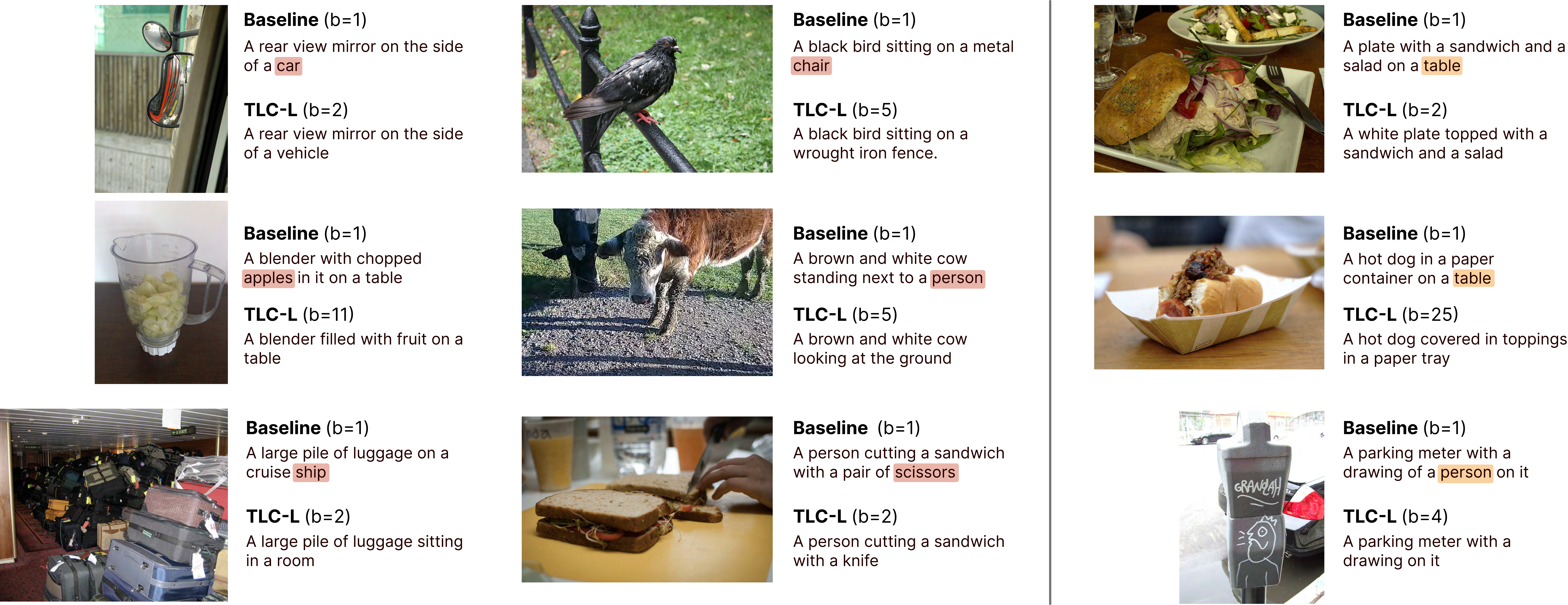}
   \caption{Qualitative examples from our test set in which \ApproachName-L avoided hallucinations in the original (Baseline) captions. In the rightmost column, we show cases where the MS~COCO object annotations did not exhaustively include all objects present. Captions are generated with \OFALarge and a beam size of 25, and $(b=i)$ refers to the index $i$ of the beam as ranked by the Baseline.
   }
   \label{fig:qual}
\end{figure*}

We now test our approach described in~\secref{sec:method-reducing-hallucinations}, where we select a caption from a set of candidates to lower the likelihood of error. We also evaluate learned confidences from \ApproachName-L, now that we can use domain-specific training data for $g$ with the image captioning validation set. Prior work~\cite{rohrbach2018emnlp} provides a framework for measuring object hallucination on MS~COCO data. \cite{rohrbach2018emnlp} provides a method to enumerate MS~COCO objects mentioned in references for a given image and enumerate objects mentioned in an arbitrary, predicted caption. We also add part-of-speech taggers~\cite{bird2009natural,spacy} to exclude predicted words that are not nouns; however, when comparing directly to prior work, we use the original implementation. A hallucination is flagged when a prediction mentioned an object not present in the reference set. This is evaluated by sentence-level and object instance-level CHAIRs and CHAIRi metrics~\cite{rohrbach2018emnlp}:

\begin{equation}
\text{CHAIR}_s = \frac{\text{\# captions with $\geq 1$ hallucination}}{\text{\# captions}}
\label{eq:chairs}
\end{equation}
\begin{equation}
    \text{CHAIR}_i = \frac{\text{\# objects hallucinated}}{\text{ \# objects mentioned}}
    \label{eq:chairi}
\end{equation}

We report standard captioning metrics~\cite{anderson2016spice,vedantam2015cider} as well as CHAIRs and CHAIRi (or CHs and CHi). We also report several caption diversity measures~\cite{shetty2017speaking,xiong2018move} to examine whether captions with lower hallucination rates reduce caption diversity: \textit{Vocab Size} measures unique unigrams across predictions, \textit{\% Novel} measures the percentage of generated captions which do not appear in the training set annotations,
\textit{Div-2} measures the ratio of unique bigrams to the number of generated words,
and \textit{Re-4} measures the repetition of four-grams.

For both \ApproachName-A and \ApproachName-L, we choose a threshold $\gamma$ on the validation set. This threshold is used at test time to make binary decisions on the correctness of a given object in a predicted caption. We extract all objects from the validation set predictions, as well as corresponding token confidences and ground-truth hallucination scores. Then, we choose a confidence level $\gamma$ that reaches at least 99\% precision when separating correct vs. hallucinated objects. This precision is intentionally very high; the OFA captioning models have fairly low rates of hallucination on MS~COCO already (as seen in~\tabref{tab:hallucination-main}), yet we are interested in pushing the caption reliability as far as possible. When aggregating token confidences over object words, we select the minimum value for \ApproachName-A and the average value for \ApproachName-L based on the validation set recall. We use a large beam size of $B = 25$ to observe the behavior of our caption selection method when given many possible candidates.

We show results from the following methods. \textbf{Standard} uses the original top caption, that is, the caption from the beam ranked highest by $f_{\mathit{cap}}$. \textbf{Standard-Aug} uses the top caption from a captioning model $f_{\mathit{cap}}'$, where its training set is augmented by the training set for $g$. This tests whether the improvements from \ApproachName-L result from using token confidence itself or from additional training data. More details on Standard-Aug are in Appendix \ref{sec:supp-model}. \textbf{ITM} uses $f_{\mathit{pre}}$ to re-rank the $B$ candidate captions from Standard based on their image-text matching score, and selects the highest-ranked caption as output. \textbf{\ApproachName-A} and \textbf{\ApproachName-L} use the respective algebraic or learned confidences over the MS~COCO object words to re-rank captions as described in \secref{sec:method-reducing-hallucinations}.

\minisection{Learned confidences lead to the least hallucinations} From~\tabref{tab:hallucination-main}, we can see that both \ApproachName-A and \ApproachName-L lower the CHs and CHi hallucination rates across all model sizes compared to the original (Standard) captions. \ApproachName-L reaches the lowest rates in each case; for example, it lowers CHs and CHi for \OFALarge by a relative 37.6\% and 34.3\% respectively. Additionally, \ApproachName-L lowers hallucination rates compared to Standard-Aug as well (\eg, a relative 20.9\% lower CHs for \OFALarge). This indicates that reserving a portion of data to train $g$ can have a bigger impact on reducing hallucinations than does using the data for augmentation. Using ITM scores slightly lessens hallucination rates over Standard, yet at the cost of large degradation in CIDEr and SPICE, and underperforms \ApproachName in all metrics. In~\tabref{tab:hallucination-beam-breakdown}, we further evaluate hallucination rates on the subset of images where the top beam from Standard was \textit{not} selected by \ApproachName-L with \OFALarge -- in other words, samples where using \ApproachName-L made a difference. This occurred in almost a quarter of the captions. Standard hallucination rates are much higher on this subset (\eg, 6.78\% CHs), whereas \ApproachName-L reduces this by at least half.

\minisection{Captioning metrics do not capture hallucinations} CIDEr and SPICE decrease across all \ApproachName-based approaches, despite having dramatic reductions in hallucination rates. This effect was also observed by~\cite{rohrbach2018emnlp}, which described how standard metrics can often fail to penalize hallucinations. For instance, the majority of a sentence might overlap with a reference caption, yet still, misclassify an object. \cite{macleod2017understanding} nevertheless find that some visually-impaired users of captioning systems prefer correctness above possibly-wrong detail, motivating the drive for low hallucination rates.

\definecolor{Gray}{gray}{0.9}
\newcolumntype{g}{>{\columncolor{Gray}}r}
\begin{table}[t]
  \centering
  \resizebox{1.\columnwidth}{!}{
  \begin{tabular}{@{\extracolsep{6pt}}llggrrrr}
    \toprule
     \multirow{2}{*}{Model} & \multirow{2}{*}{Confidence} & \multicolumn{2}{c}{Hallucination} & \multicolumn{2}{c}{Quality} \\ \cline{3-4} \cline{5-6} 
     
    \rule{0pt}{2.5ex} & & \multicolumn{1}{c}{CHs $(\downarrow)$ }  & \multicolumn{1}{c}{CHi $(\downarrow)$} & \multicolumn{1}{c}{CIDEr $(\uparrow)$} & \multicolumn{1}{c}{SPICE $(\uparrow)$} \\
    \midrule
 \multirow{5}{*}{\OFALarge} & Standard-Aug & 2.20 & 1.38 & 153.3 & 26.7 \\ \cmidrule{2-6}
     \rule{0pt}{1.2ex}  & Standard & 2.79 & 1.78 & \textbf{144.4} & \textbf{25.8} \\
    & ITM & 2.57 & 1.76 & 126.5 & 24.4 \\
    & \ApproachName-A & 1.81 & 1.24 & 140.7 & 25.5 \\
    & \ApproachName-L & \textbf{1.74} & \textbf{1.17} & 141.8 & 25.4  \\
    \midrule
     \multirow{5}{*}{\OFABase}  & Standard-Aug & 3.00 & 1.89 & 148.8 & 26.1  \\ \cmidrule{2-6}
     \rule{0pt}{1.2ex} & Standard & 3.78 & 2.39 & \textbf{142.9} & \textbf{25.6} \\
     & ITM & 3.22  &  2.15  & 127.1 & 24.3  \\
    & \ApproachName-A & 2.47 & 1.75 & 137.5 & 25.2 \\
     & \ApproachName-L & \textbf{2.05} &\textbf{1.48} & 137.5 & 24.9  \\
     \midrule
      \multirow{5}{*}{\OFATiny} & Standard-Aug & 10.58 & 6.83 & 119.8 & 22.1 \\ \cmidrule{2-6}
     \rule{0pt}{1.2ex} &  Standard & 11.01  & 7.23 & \textbf{117.4} & \textbf{21.7}   \\
      & ITM & 9.42 & 6.51 & 106.6 & 20.6  \\
    & \ApproachName-A & 9.87 & 6.86 & 115.8 & 21.5 \\
      & \ApproachName-L & \textbf{8.79} & \textbf{6.43} & 113.9 & 21.3 \\
    \bottomrule
  \end{tabular}
  \caption{Hallucination rates and captioning metrics on our test set when generating captions with a beam size of 25.
  }
  \label{tab:hallucination-main}}
\end{table}

\definecolor{Gray}{gray}{0.9}
\begin{table}[t]
  \centering
  \resizebox{1.\columnwidth}{!}{
  \begin{tabular}{@{}llrrrr}
    \toprule
    \multirow{2}{*}{Model} & \multirow{2}{*}{Conf.} & \multicolumn{1}{c}{Vocab} &\multicolumn{1}{c}{\% Novel} & \multicolumn{1}{c}{Div-2} & \multicolumn{1}{c}{Re-4} \\
     & &  \multicolumn{1}{c}{Size $(\uparrow)$} & \multicolumn{1}{c}{$(\uparrow)$} &  \multicolumn{1}{c}{$(\uparrow)$} & \multicolumn{1}{c}{$(\downarrow)$} \\
    \midrule
 \multirow{3}{*}{\OFALarge} & Std. & 2822 & 77.07  & 6.97 & 66.34 \\
    & \ApproachName-A &  \textbf{2980} & \textbf{78.97} & \textbf{7.37} & \textbf{64.74} \\
    & \ApproachName-L & \underline{2915} & \underline{77.70}  & \underline{7.13} & \underline{65.54} \\
    \midrule
     \multirow{3}{*}{\OFABase}  & Std. & 2272 & 75.43  & 5.68 & 71.14 \\
    & \ApproachName-A & \textbf{2453} & \textbf{78.49} & \textbf{6.13} & \underline{69.28} \\
     & \ApproachName-L & \underline{2452} & \underline{77.53}  & \underline{6.03} & \textbf{69.76} \\
     \midrule
      \multirow{3}{*}{\OFATiny} & Std. &  1130 & 74.80  & 2.73 & 83.29  \\
    & \ApproachName-A & \underline{1211} & \underline{75.71}  & \underline{2.91} & \underline{82.68} \\
      & \ApproachName-L & \textbf{1243} & \textbf{77.05} & \textbf{3.01} & \textbf{82.12}  \\
    \bottomrule
  \end{tabular}
  \caption{Caption diversity metrics, evaluated on our test set.}
  \label{tab:hallucination-diversity}}
\end{table}

\begin{table}
    \centering
    \resizebox{1\columnwidth}{!}{
    \begin{tabular}{@{}lrrrr}
    \toprule
        Subset  & $\#$ I & Method & CHs ($\downarrow$) & CHi ($\downarrow$)   \\
        \midrule
         \multirow{2}{*}{Full test set} &  \multirow{2}{*}{20,252} & Standard & 2.79 & 1.78  \\
         & & \ApproachName-L & \textbf{1.74} & \textbf{1.17}  \\
       \midrule
    \multirow{2}{*}{\ApproachName-L, $b > 1$} &  \multirow{2}{*}{5,401} & Standard & 6.78  & 3.22  \\
    & & \ApproachName-L & \textbf{2.81} & \textbf{1.61} \\
         \bottomrule
    \end{tabular}
    }
    \caption{Top: Results on the full test set reported in~\tabref{tab:hallucination-main}. Bottom: Hallucination rates on a subset of images where \ApproachName-L did not choose the top beam. $\#$ I denotes the number of images in each set. Results are shown for \OFALarge.
 }
    \label{tab:hallucination-beam-breakdown}
\end{table}

\minisection{\ApproachName improves caption diversity} From~\tabref{tab:hallucination-diversity}, our method achieves higher performance on diversity metrics across all model sizes. For instance, \ApproachName-A consistently increases bigram uniqueness score \textit{Div-2}, and decreases the repetition measure \textit{Re-4}. Incorporating confidence into caption selection may help overcome language priors, where co-occurrence statistics from training influence token likelihoods. Diversity can improve as a result, where captions are driven more by consistency with the image rather than language. For example, the top center sample in~\figref{fig:qual} shows the baseline hallucinating a ``metal chair'', compared to the correct yet uncommon words ``wrought iron fence'' described by \ApproachName-L.

\definecolor{Gray}{gray}{0.9}
\newcolumntype{g}{>{\columncolor{Gray}}r}
\begin{table*}[t!]
     \resizebox{\textwidth}{!}{
    \begin{tabular}{@{}ll|c|rrrr@{}r@{\ \ }r|rrrr@{\ }r@{\ \ }r@{}}
    \toprule
Reported in  & Method & Beam & \multicolumn{6}{c|}{XE Loss} & \multicolumn{6}{c}{SC Loss} \\
      &        & Size & \multicolumn{1}{c}{B@4} & \multicolumn{1}{c}{S} & \multicolumn{1}{c}{M} & \multicolumn{1}{c}{C} & \multicolumn{1}{c}{CHs ($\downarrow$)} & CHi ($\downarrow$) & \multicolumn{1}{c}{B@4} & \multicolumn{1}{c}{S} & \multicolumn{1}{c}{M} & \multicolumn{1}{c}{C} & CHs ($\downarrow$) & CHi ($\downarrow$) \\
        \midrule
\cite{rohrbach2018emnlp} EMNLP 2018 & NBT \cite{lu2018neural} & 5 & - & 19.4 & 26.2 & 105.1 & 7.4 & 5.4 & - & - & - & - & - & - \\
\cite{rohrbach2018emnlp} EMNLP 2018 & TopDown \cite{anderson2018bottom} (no Boxes) & 5 & - & 19.9 & 26.7 & 107.6 & 8.4 & 6.1 & - & 20.4 & 27.0 & 117.2 & 13.6 & 8.8 \\
\cite{rohrbach2018emnlp} EMNLP 2018 & TopDown \cite{anderson2018bottom} & 5 & - & 20.4 & 27.1 & 113.7 & 8.3 & 5.9 & - & 21.4 & 27.7 & 120.6 & 10.4 & 6.9 \\
\cite{yang2021causal} CVPR 2021 & Transformer & unk & - & - & - & - & - & - & 38.6 & 22.0 & 28.5 & 128.5 & 12.1 & 8.1 \\
\cite{yang2021causal} CVPR 2021 & Transformer+CATT & unk & - & - & - & - & - & - & 39.4 & 22.8 & 29.3 & 131.7 & 9.7 & 6.5 \\
\cite{yang2021deconfounded} PAMI 2021 & UD-DICv1.0 & 5 & - & - & - & - & - & - & 38.7 & 21.9 & 28.4 & 128.2 & 10.2 & 6.7 \\
\cite{biten2022let} WACV 2022 & UD-L & no & 34.4 & 20.7 & 27.3 & 112.7 & 6.4 & 4.1 & 37.7 & 22.1 & 28.6 & 124.7 & 5.9 & 3.7 \\
\cite{biten2022let} WACV 2022 & UD-L + Occ & no & 33.9 & 20.3 & 27.0 & 110.7 & 5.9 & 3.8 & 37.7 & 22.2 & 28.7 & 125.2 & 5.8 & 3.7 \\
\cite{liu2022show} CVPR 2022 & CIIC$_G$ & 3 & 37.3 & 21.5 & 28.5 & 119.0 & 5.3 & 3.6 & 40.2 & 23.2 & 29.5 & 133.1 & 7.7 & 4.5 \\
\cite{li2022comprehending} CVPR 2022 & COS-Net & 3 & 39.1 & 22.7 & 29.7 & 127.4 & 4.7 & 3.2 & 42.0 & 24.6 & 30.6 & 141.1 & 6.8 & 4.2 \\

This work & OFA$_{\text{Large}}$ \cite{ofa} & 5 & \textbf{41.8} &\textbf{ 24.4} & \textbf{31.3} & \textbf{140.7} & 3.1 & 2.0 & \textbf{42.3 }& \textbf{25.5} &\textbf{ 31.6} & \textbf{145.0 }& 3.1 & 2.0 \\
This work   & OFA$_{\text{Large}}$  + \ApproachName-L & 5 & 41.2 & 24.1 & 30.9 & 138.4 & \textbf{*2.0}& \textbf{*1.4} & 42.0 & 25.2 & 31.4 & 143.8 & \textbf{2.3} & \textbf{1.5} \\
         
         \bottomrule
    \end{tabular}
    }
    \caption{Comparison to prior work for hallucination in image captioning on the MS~COCO Karpathy test split. Although we add a noun parser for our results in Tables \ref{tab:hallucination-main}, \ref{tab:hallucination-diversity}, and \ref{tab:hallucination-beam-breakdown}, we remove this step here and use the original evaluation provided by~\cite{rohrbach2018emnlp} to be consistent with prior work. We show captioning metrics B@4 (BLEU~\cite{papineni2002bleu}), S (SPICE~\cite{anderson2016spice}), M (METEOR~\cite{lavie2007meteor}), and C (CIDEr~\cite{vedantam2015cider}). \textbf{*} indicates state-of-the-art for hallucination rates. }
    \label{tab:hallucination-prior}
\end{table*}

\minisection{Qualitative analysis} We show several qualitative examples in~\figref{fig:qual}.
In the left column, we see two examples where \ApproachName-L ``backed-off'' to a more general concept, whereas the baseline was specific, yet the image did not contain enough information to determine whether the specificity was indeed correct (\eg, ``car'' vs. ``vehicle'' and ``apples'' vs. ``fruit'').
A prior work~\cite{guadarrama2013youtube2text} explicitly optimized for this hierarchical generalization of unknown concepts, whereas here it emerges when considering confidence.
\ApproachName-L also avoids misclassification errors, such as ``person'' or ``scissors'' in the middle column. On the right column, we show examples influenced by incomplete object annotations. For example, the reference segmentations and captions might overlook the object ``table''. \ApproachName-L rejects captions that mention ``table'' in some of these cases, reflecting its training objective where correctness was judged based on faithfulness to the reference distribution. We include additional examples, including several failure cases, in Appendix~\ref{sec:supp-qual}.

\minisection{\ApproachName-L with \OFALarge sets a new state-of-the-art} We compare to previous results on MS~COCO object hallucination in~\tabref{tab:hallucination-prior}. We re-train our captioning models and confidence estimators on a dataset split that does not overlap with the Karpathy test split used for evaluation~\cite{karpathy2015deep}. \cite{rohrbach2018emnlp} show that training with a self-critical (SC) loss after training with cross-entropy (XE)~\cite{rennie2017self} can improve captioning metrics, yet worsen hallucination rates compared to training with XE alone.
We find that the baseline \OFALarge has similar hallucination rates for XE and SC, yet \ApproachName-L indeed produces the least hallucinations on top of the XE model.
This leads to a new state-of-the-art of 2.0\% and 1.4\% for CHs and CHi respectively. Notably, \ApproachName-L reduces hallucination without requiring any architecture changes to its captioning model, in contrast to the prior SOTA of COS-Net, where specific modules were introduced to capture image semantics.

\section{Discussion and Limitations}
\label{sec:limitations}

While \ApproachName-L provides effective confidence estimates for caption generation, it requires domain-specific training data for learning a confidence estimator from scratch on top of captioning model features. \ApproachName-A, on the other hand, uses the captioning model outputs directly, which leverages generalization ability from large-scale pretraining. Thus, \ApproachName-A can be effectively applied in settings where in-domain training data for captioning is not available. To combine these advantages, future research could explore unsupervised methods for learning correctness. Additionally, we use algebraic confidence estimates from uncalibrated output distributions, where output probabilities do not necessarily match actual probabilities of correctness. Potential future work may apply calibration methods to token-level confidence for improving caption correctness. Finally, learned confidences may also be incorporated into decoding methods that are not autoregressive.

\section{Conclusion}
\label{sec:conclusion}

In this work, we have explored a simple method using \ApproachNameLong (\ApproachName) for determining whether a caption correctly describes an image, a critical part of vision-language understanding. 
We find that judging caption correctness at a finer granularity than existing approaches leads to improvements in several settings, such as evaluating compositional reasoning with image-caption pairs or reducing object hallucinations in generated captions. To do so, \ApproachName uses a vision-language model fine-tuned on image captioning to produce token confidences, and then aggregates either algebraic (\ApproachName-A) or learned token confidences (\ApproachName-L) over words or sequences to estimate image-caption consistency. Increasing the confidence granularity with \ApproachName-A improves over prior state-of-the-art image and group scores on Winoground~\cite{thrush_and_ross2022winoground} by a relative 37\% and 9\%, respectively, and improves accuracy in verb understanding on SVO-Probes~\cite{hendricks2021probing} by a relative 10\%. When training data are available to learn and calibrate confidences with \ApproachName-L, we reduce object hallucination rates on COCO Captions by a relative 30\%, setting a new state-of-the-art. Overall, our results demonstrate that token-level confidence, whether algebraic or learned, can be a powerful yet simple resource for reducing errors in captioning output and assessing image-caption consistency.

\minisection{Acknowledgements} We thank David Chan, Kate Saenko, and Anastasios Angelopoulos for helpful discussions and feedback.  Authors, as part of their affiliation with UC Berkeley, were supported in part by the NSF CISE Expeditions Award CCF-1730628, DoD, including DARPA’s LwLL, PTG, and/or SemaFor programs, the Berkeley Artificial Intelligence Research (BAIR) industrial alliance program, as well as gifts from Astronomer, Google, IBM, Intel, Lacework, Microsoft, Mohamed Bin Zayed University of Artificial Intelligence, Nexla, Samsung SDS, Uber, and VMware.

{\small
\bibliographystyle{ieee_fullname}
\bibliography{egbib}
}

\clearpage

\appendix

\section*{Appendix}

\section{Overview}
\label{sec:supp-overview}

\cref{sec:supp-precision-recall} presents an ablation showing several alternative algebraic confidence estimates, and compares the precision-recall curve for the learned \ApproachName-L to that of algebraic confidences when separating correct and hallucinated objects.
\cref{sec:supp-qual} presents additional qualitative examples of both success and failure cases, comparing \ApproachName-L to the Baseline model. \cref{sec:supp-dataset} and \cref{sec:supp-model} provide further details on datasets and models respectively.

\section{Alternative Confidence Estimates}
\label{sec:supp-precision-recall}

\begin{figure*}[h]
  \centering
   \includegraphics[width=\linewidth]{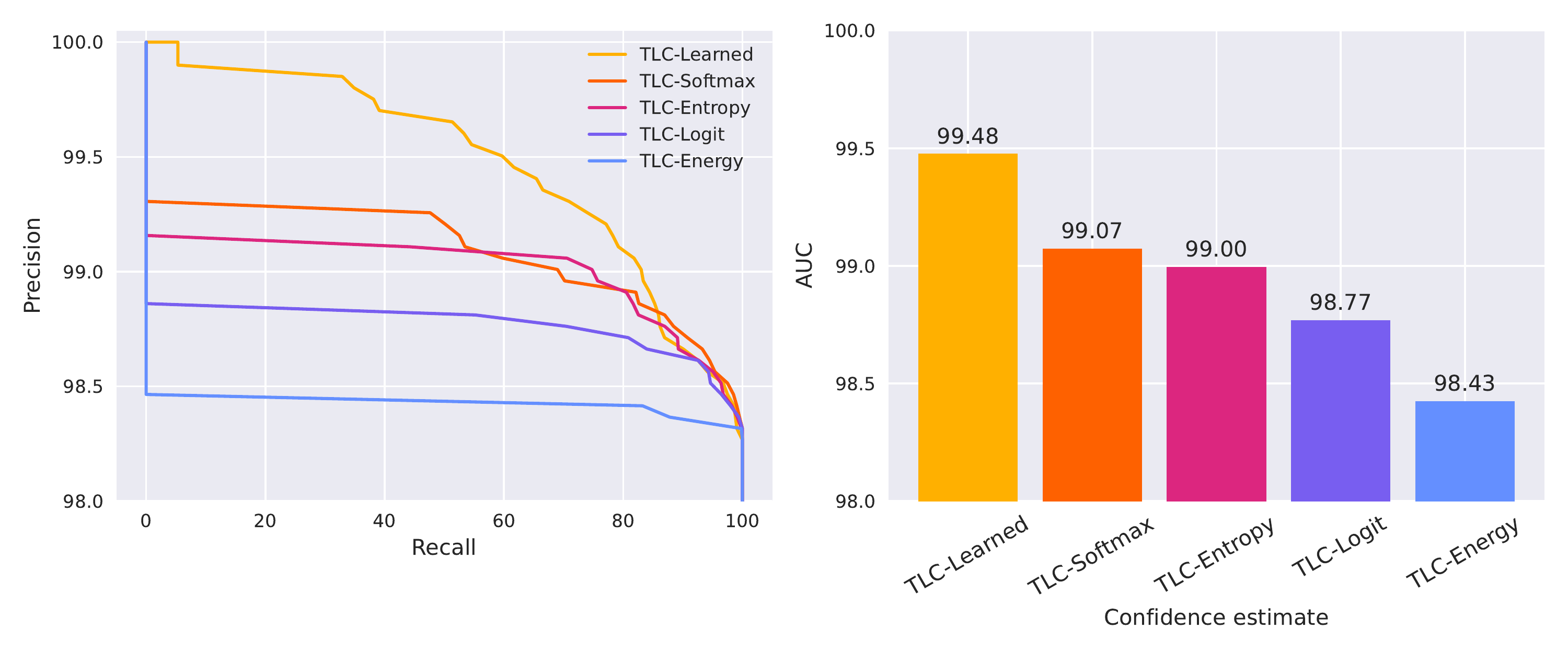}
   \caption{Precision-recall curve (left) and AUC (right) with different confidence estimates for separating correct and hallucinated objects. Results are shown on our validation set using \OFALarge.
   }
   \label{fig:supp-pr}
\end{figure*}

We compare several other choices of algebraic confidence estimates for \ApproachName-A besides softmax score used in the main paper. All are derived from the likelihood (logit) distribution $\vec z_k$, as mentioned in \secref{sec:method-algebraic}. \textbf{Logit} is the logit value for the selected token directly from $\vec z_k$, whereas \textbf{Softmax} is the corresponding value after a softmax function. Again, in our main paper, \ApproachName-A is based on this softmax score confidence.
\textbf{Entropy} is the negative entropy of the log-softmax distribution, as a higher entropy should indicate higher uncertainty. Entropy has been previously used as a direct estimate of model uncertainty~\cite{wang2020tent} as well as a penalty in image caption decoding~\cite{xiao2021hallucination}. Finally, we consider the \textbf{Energy} score~\cite{liu2020energy}, originally proposed as a measure for OOD detection that theoretically correlates with the probability density of the in-domain samples. We use a temperature of 1, and negate the energy score so positive values indicate confident samples. 

In \figref{fig:supp-pr}, we show the precision-recall curve for various confidence estimates to separate correct and hallucinated objects. We compute these results on our MSC-Main validation set for $g$ (see \tabref{tab:supp-datasets}). Specifically, we are not interested in the exact values of confidence estimates themselves, but rather how well they can \textit{rank} correct objects over those that are hallucinated. When using confidence estimates in practice, we need a threshold to make a binary decision about whether an object in a caption is considered hallucinated or not (\secref{sec:method-reducing-hallucinations}). We choose this threshold for a specific precision level, above the accuracy that the model achieves on its own. For instance, on the validation set for $g$, about 98.3\% of the captioning model's predicted objects are correct (and the rest hallucinated). To push reliability further, we choose a threshold $\gamma$ for each method that achieves a precision of 99\%. In \figref{fig:supp-pr} (left), we therefore only show recall rates above 98\% precision, yet show the overall area-under-the-curve (AUC) in \figref{fig:supp-pr} (right).

From \figref{fig:supp-pr}, we can see that \ApproachName-Learned (\ie, \ApproachName-L) achieves the highest AUC of 99.48\%, and \ApproachName-Softmax achieves the second-highest of 99.07\%. The precision-recall plot shows that all algebraic confidences reach 0\% recall before 99.5\% precision, whereas \ApproachName-L still retains about 60\% recall at this high precision rate. In our main paper, we use \ApproachName-A to denote \ApproachName-Softmax, as it performed the best among the algebraic confidences.

\section{Additional Qualitative Examples}
\label{sec:supp-qual}

\begin{figure*}
  \centering
     \includegraphics[width=\linewidth]{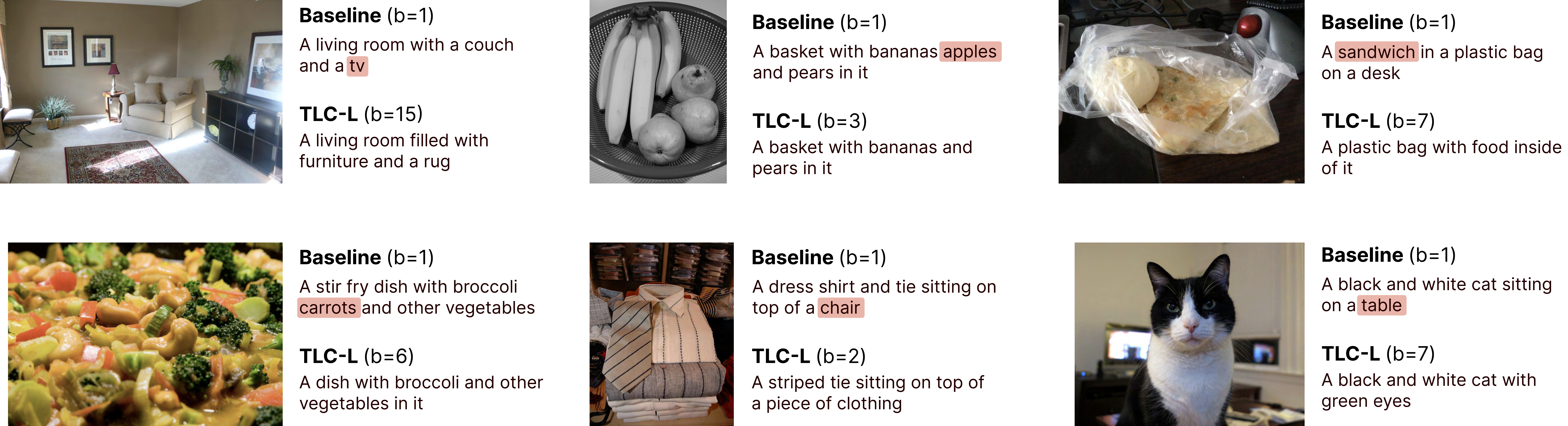}
  \caption{Additional qualitative examples on our test set for \ApproachName-L on \OFALarge, where the Baseline model caption contained a hallucination, yet the caption selected by \ApproachName-L did not.
  }
  \label{fig:supp-qual-success}
\end{figure*}
\begin{figure*}
  \centering
     \includegraphics[width=\linewidth]{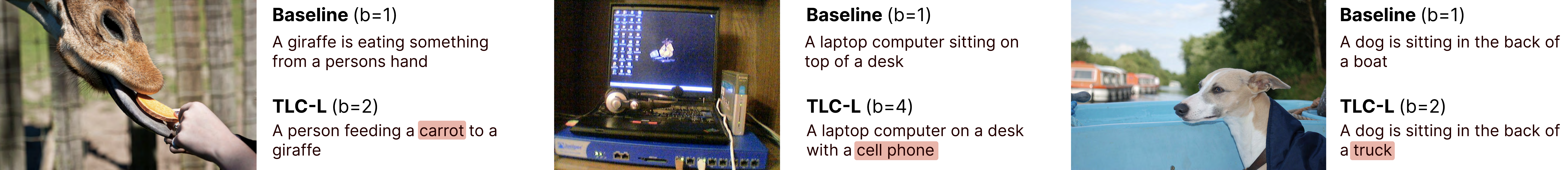}
  \caption{Failure cases on our test set for \ApproachName-L on \OFALarge, where \ApproachName-L selected a caption with a hallucination, yet the Baseline did not.
  }
  \label{fig:supp-qual-failure}
\end{figure*}

In \figref{fig:supp-qual-success}, we present qualitative examples (in addition to those in \figref{fig:qual}) where the Baseline model caption contained a hallucination, yet the caption selected by \ApproachName-L did not. Note that ``Baseline'' refers to ``Standard'' as in \tabref{tab:hallucination-main}. In \figref{fig:supp-qual-failure}, we show several failure cases of \ApproachName-L. On the left is a case where the Baseline model selects a more general caption, whereas \ApproachName-L erroneously rejects it for one with a hallucinated ``carrot''. On the middle and right, \ApproachName-L selects captions that include other hallucinations of objects. Nevertheless, \ApproachName-L corrected 44.5\% ($252/566$) of captions that contained a hallucination from the Baseline model, whereas \ApproachName-L introduced a hallucination in only 0.2\% ($38/19,686$) of captions that did not contain a hallucination from the Baseline model.

\section{Dataset details}
\label{sec:supp-dataset}

\minisection{MS~COCO Captions} We use the same dataset splits as~\cite{whitehead2022reliable} for training and validating the captioning model $f_{\mathit{cap}}$ and confidence estimator $g$, as~\cite{whitehead2022reliable} similarly reserves validation data in MS~COCO for training a confidence estimator (yet for the visual question answering task, rather than image captioning). For the Standard-Aug model $f_{\mathit{cap}}'$ in \tabref{tab:hallucination-main}, we include the training set for $g$ as part of the training set for $f_{\mathit{cap}}'$.  In~\tabref{tab:supp-datasets}, we refer to these splits as MSC-Main (for MS~COCO Main), and use them for results in Tabs. \ref{tab:hallucination-main}, \ref{tab:hallucination-diversity}, and \ref{tab:hallucination-beam-breakdown}, and Figs. \ref{fig:qual}, \ref{fig:supp-pr}, \ref{fig:supp-qual-success}, and \ref{fig:supp-qual-failure}. For comparison to prior work that uses the Karpathy test split (\tabref{tab:hallucination-prior}), we re-split the validation set to prevent overlap. These details are presented as MSC-Prior in~\tabref{tab:supp-datasets}.

\begin{table}
  \centering
  \resizebox{1.\columnwidth}{!}{
  \begin{tabular}{l@{\hskip12pt}lrr}
    \toprule
    Dataset & Use Case & \# Images & \# Captions \\

\midrule

\multirow{4}{*}{MSC - Main } & Train $f_{\mathit{cap}}$ and $f_{\mathit{cap}}'$ & 82,783 & 414,113 \\
&  Validate $f_{\mathit{cap}}$, Train $g$ and $f_{\mathit{cap}}'$  & 16,202 & 81,065 \\
& Validate $g$ and $f_{\mathit{cap}}'$, Select $g$ thresholds & 4,050 & 20,268  \\
& Evaluation  & 20,252 & 101,321 \\ 
\midrule

\multirow{4}{*}{MSC - Prior} & Train $f_{\mathit{cap}}$ & 82,783 & 414,113  \\
& Validate $f_{\mathit{cap}}$, Train $g$ & 28,403 & 142,120  \\
& Validate $g$, Select $g$ thresholds &  7,101 & 35,524  \\
& Evaluation  & 5,000 & 25,010  \\ 
\midrule

Winoground & Evaluation &  800 & 800 \\
\midrule
SVO-Probes & Evaluation & 12,958 & 6,479 \\

\bottomrule
  \end{tabular}
  \caption{Overview of datasets used in our work. MSC indicates MS~COCO Captions~\cite{chen2015microsoft}. }
  \label{tab:supp-datasets}
  }

\end{table}

\minisection{Winoground} We use the original data and evaluation setup for Winoground as in the original paper~\cite{thrush_and_ross2022winoground}, which consisted of 800 unique images and captions. This leads to 400 examples, each consisting of two image-caption pairs, where the captions contain the same words and/or morphemes yet a different word order.

\minisection{SVO-Probes} For SVO-Probes~\cite{hendricks2021probing}, we use the authors' public code to access a subset of data where the images were available. As discussed in \secref{sec:experiment-svo}, each image is annotated with a  $\langle$subject, verb, object$\rangle$ relation, \eg,  $\langle$girl, sit, shore$\rangle$ relation. We take the available data that contrasts two verbs, \eg, a ``positive'' or image-consistent relation $\langle$girl, sit, shore$\rangle$ and a ``negative'' or inconsistent relation $\langle$girl, walk, shore$\rangle$. For each image, we take the provided ``positive'' caption (\eg, ``A girl sits on the shore''), and use a part-of-speech tagger~\cite{spacy} to localize the verb (\eg, ``sit'') in the sentence. We do not use images where the tagger failed to identify the verb, often in cases where the verb did not appear in the caption itself (\eg,  a triplet of $\langle$person, wear, glasses$\rangle$ with a caption of ``The glasses fogged up''). The final split contains about 6,500 image-caption pairs (\tabref{tab:supp-datasets}), half of which are correct pairs. This evaluation is not directly comparable to prior work~\cite{hendricks2021probing}, which used the full set of data, chose a threshold of 0.5 to indicate whether or not an individual sample matched an image, and was performed at a sequence-level rather than word-level. In our work, we contrast a positive and negative image for a given caption, and label a sample as correct if the confidence for the positive pair is larger than the confidence for the negative pair, similar to Winoground.

\minisection{Overlap with training data} All OFA models were not exposed to any MS~COCO validation or test data during pretraining~\cite{ofa}. Winoground was hand-curated from the Getty Images API~\cite{getty,thrush_and_ross2022winoground}, which is not used by OFA pretraining. Data from SVO-Probes was collected via the Google Image Search API and de-duplicated against Conceptual Captions~\cite{hendricks2021probing,sharma2018conceptual}. As OFA models used Conceptual Captions during pretraining, we assume there is no further overlap. 

\section{Model details}
\label{sec:supp-model}

\minisection{Captioning} To complement the details in \secref{sec:experiment-setup}, we provide additional experimental details for the captioning models. We use publicly available checkpoints for pretrained models provided by~\cite{ofa}. Parameter counts are 930M for \OFALarge, 180M for \OFABase, and 33M for \OFATiny~\cite{ofa}. To finetune the pretrained models on MS~COCO Captions, we follow the same settings from~\cite{ofa}, where we train with cross entropy loss for 2 epochs for \OFALarge, and 5 epochs for \OFABase and \OFATiny. We then train with CIDEr optimization for 3 epochs.

\minisection{\ApproachName-L} In addition to details in \secref{sec:experiment-setup}, we provide further information on the learned confidence estimator $g$. We use a 4-layer Transformer encoder~\cite{vaswani2017attention} with 4 attention heads each. The embedded output corresponding to the token of interest $t_k$ (\secref{sec:method-learning}) is passed to a 2-layer MLP, with hidden dimensions of size 512. The embedding dimension is 1024 for \OFALarge, 768 for \OFABase, and 512 for \OFATiny.  We train $g$ for 200 epochs, with a batch size of 256, starting learning rate of 0.001, warm up ratio of 0.06 and polynomial learning rate decay to 2$e$-7. We use the Adam optimizer~\cite{kingma2014adam} and clip gradients over 1.0. For aggregating tokens over objects for caption generation (\secref{sec:method-reducing-hallucinations}), we use the minimum score for softmax and average for \ApproachName-L, found on our validation set.

\end{document}